\documentclass{article}
\usepackage{authblk}

\usepackage{hyperref}

\usepackage{mathptmx}       
\usepackage{helvet}         
\usepackage{courier}        
\usepackage{type1cm}        
\usepackage{float}                            
\usepackage{makeidx}         
\usepackage{graphicx}        
                             \usepackage{multicol}        
\usepackage[bottom]{footmisc}
\usepackage{graphicx}      

\usepackage{graphics} 
\usepackage{graphicx}
\usepackage[table]{xcolor}
\usepackage{caption}
\usepackage{algpseudocode}
\usepackage{algorithm}
\usepackage[]{units}
\usepackage{paralist}
\usepackage{mathtools}

\usepackage{xcolor}
\usepackage{siunitx}
\usepackage{multirow}
\usepackage{epstopdf}
\usepackage{epstopdf}
\graphicspath{ {Pictures/} }
\usepackage[export]{adjustbox}
\usepackage{tikz}
\usetikzlibrary{arrows}
\usetikzlibrary{calc}
\usetikzlibrary{backgrounds}
\usetikzlibrary{decorations.footprints}
\usetikzlibrary{decorations.pathmorphing}
\usetikzlibrary{decorations.text}
\usetikzlibrary{fadings}
\usetikzlibrary{fit}
\usetikzlibrary{matrix}
\usetikzlibrary{mindmap}
\usetikzlibrary{patterns}
\usetikzlibrary{positioning}
\usetikzlibrary{shadows}
\usetikzlibrary{shapes}
\usetikzlibrary{trees}
\usetikzlibrary{external}

\usepackage{epsfig}
\usepackage{todonotes}
\usepackage[]{units}
\usepackage{balance}
\usepackage{float}
\usepackage{caption}
\usepackage{amsmath} 
\usepackage{amssymb}  
\usepackage{mathtools}
\graphicspath{ {Pictures/} }
\DeclareGraphicsExtensions{.pdf,.eps}

\def\antenna#1#2#3#4{\begin{scope}[shift={#1}, rotate=#2, scale=#3]
  \draw[line width=1.5,line cap=round,#4] (0,0.03) -- (0,0.1);
  \draw[line width=1,line cap=round,#4] (0,0.1) -- (0,0.17);
  \draw[line width=1,line cap=round,#4] (0,0.1) -- (-0.05,0.15);
  \draw[line width=1,line cap=round,#4] (0,0.1) -- (0.05,0.15);
\end{scope}}

\def\TablesColumnsColor{black!4}

\newcolumntype
{g}
{
	>{\centering \columncolor{\TablesColumnsColor} \arraybackslash}
	p{0.15\textwidth}
	<{}
}

\newcolumntype
{w}
{
	>{\centering \arraybackslash}
	p{0.15\textwidth}
	<{}
}

\title{Autonomous visual inspection of large-scale infrastructures using aerial robots}

\author[]{Christoforos Kanellakis*}
\author[]{Emil Fresk*}
\author[]{Sina Sharif Mansouri*}
\author[]{Dariusz Kominiak}
\author[]{George Nikolakopoulos}
\affil[]{Robotics Team, Department of Computer Science, Space and Electrical Engineering,\\ Lule\r{a} University of Technology, Lule\r{a} SE-97187, Sweden\\ 
\texttt{{Emails} \{chrkan,emifre,sinsha,darkom,geonik\}@ltu.se}\\
* Authors contributed equally}
\date{}

\begin{document}
\maketitle

\begin{abstract}
This article presents a novel framework for performing visual inspection around 3D infrastructures, by establishing a team of fully autonomous Micro Aerial Vehicles (MAVs) with robust localization, planning and perception capabilities. The proposed aerial inspection system reaches high level of autonomy on a large scale, while pushing to the boundaries the real life deployment of aerial robotics. In the presented approach, the MAVs deployed for the inspection of the structure rely only on their onboard computer and sensory systems. The developed framework envisions a modular system, combining open research challenges in the fields of localization, path planning and mapping, with an overall capability for a fast on site deployment and a reduced execution time that can repeatably perform the inspection mission according to the operator needs. The architecture of the established system includes: 1) a geometry-based path planner for coverage of complex structures by multiple MAVs, 2) an accurate yet flexible localization component, which provides an accurate pose estimation for the MAVs by utilizing an Ultra Wideband fused inertial estimation scheme, and 3) visual data post-processing scheme for the 3D model building. The performance of the proposed framework has been experimentally demonstrated in multiple realistic outdoor field trials, all focusing on the challenging structure of a wind turbine as the main test case. The successful experimental results, depict the merits of the proposed autonomous navigation system as the enabling technology towards aerial robotic inspectors. 
\end{abstract}


%

\section{Background \& Related Works}
Nowadays, Micro Aerial Vehicles (MAVs) are gaining more and more attention from the scientific community, constituting a fast-paced emerging technology that constantly pushes their limits for accomplishing complex tasks~\cite{Kanellakis2017}. These platforms are characterized by their mechanical simplicity, agility, stability and outstanding autonomy to reach remote and distant places. Endowing MAVs with proper sensor suites, while navigating in indoors/outdoors, cluttered and complex environments, could establish them as a powerful aerial tool for a wide span of applications. Some characteristic examples of application scenarios for such a novel deployment of the aerial technology include infrastructure inspection~\cite{mansouri2018cooperative, TORRES2016441}, public safety-surveillance~\cite{michael2012collaborative}, and search and rescue missions~\cite{tomic2012toward}.

One of the most common application areas that MAVs are employed, is in the filming industry, but there are efforts from other industries such as Mining, Oil, and Energy Providers, to invest in the commercialization of MAVs to perform remote inspection applications. Towards this vision, MAVs are powerful tools that have the profund potential to decrease the risks of human life, decrease the execution time and increase the efficiency of the overall inspection task, especially when compared to conventional methods~\cite{sesar2016european}. Despite the fact that the research in the aerial robotics has reached significant milestones regarding localization~\cite{perez2018architecture}, planning~\cite{achtelik2014motion} and perception~\cite{scaramuzza2014vision, GARCIAPULIDO2017152}, successful real-life demonstrations of autonomous inspection systems have been rarely reported in the literature, with the majority of the applications focusing on impressive laboratory trials under full control environments and in most of the cases under the utilization of expensive motion capturing systems~\cite{lupashin2014platform} or small scale and well defined outdoor environments~\cite{teixeira2017real,forster2015continuous}. 

\begin{figure}[t]
\centering
\includegraphics[width = 0.8\linewidth]{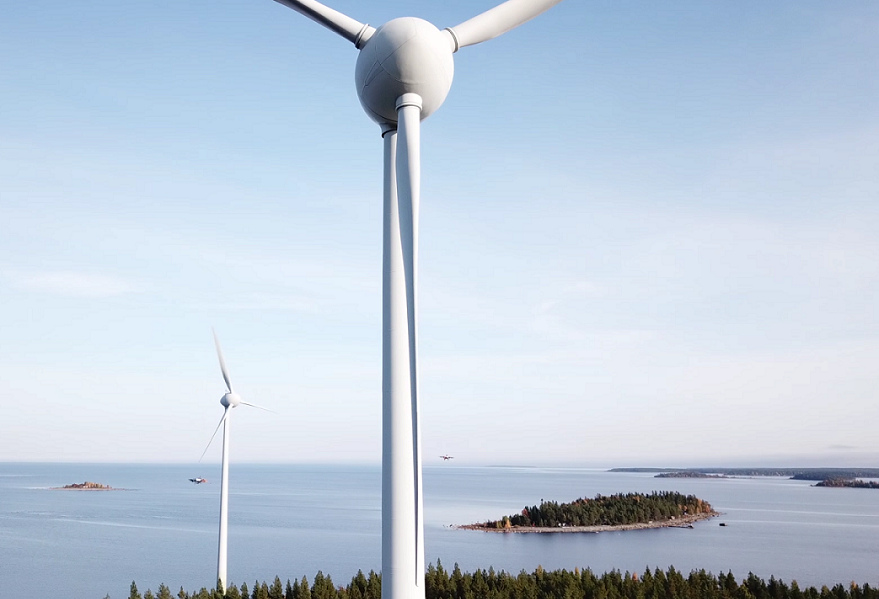}
\caption{The main inspected wind turbine and the MAV, left to the wind turbine in the background, while flying around the blades, where it should be noted the profound scale of the infrastructure in relation to the MAV.}
\label{inspection_concept}
\end{figure}

Thus, one of the most important contributions of this article is the establishment of an aerial system capable to visually pre-inspect fully autonomously an outdoors large scale infrastructure, through the coordination of multiple aerial vehicles. Towards this contribution, the article will further contribute with the implementation of a novel and accurate localization enabled scheme for collaborative aerial inspection of infrastructure, a scheme that is based on Ultra WideBand (UWB) distance measurements and Inertial Measurement Unit (IMU) sensor fusion. In this approach, the aerial platforms navigate autonomously based on the UWB-Inertial fused state estimation, using a local UWB network, placed around the structure of inspection. A second contribution of this article is the experimental evaluation of a Collaborative Coverage Path Planner (C-CPP) algorithm that has the ability to guarantee the full coverage of the infrastructure by considering camera, geometry, collision, and other application posed constraints. The coverage path is generated for every MAV, based on the structure geometric characteristics, while identifying and assigning parts of the structure to different agents, leading to faster inspection times. As an outcome, the covered path guarantees an overlapping Field of View (FoV) to enable the generation of an off-line 3D model of the structure. The final contribution stems from the real life successful demonstration of a fully functional on-board visual sensor scheme that it is able to have the dual role of providing: a) low resolution compressed data for the visual assessment of the structure, and b) high resolution for post processing e.g. build 3D models and area image stitching. The overall concept of the proposed collaborative aerial inspection scheme is presented in Figure~\ref{inspection_concept}, where two collaborative MAVs are performing an aerial inspection of a wind turbine with a corresponding video~\footnote{\url{https://youtu.be/z_Lu8HvJNoc}}.

The rest of the article is structured as it follows. The overall system is described in Section~\ref{Aerial Inspection System}. More specifically, Section~\ref{Cooperative Coverage Path Planner} presents the novel geometric approach for the C-CPP problem for infrastructure inspection. Section~\ref{Inertial Odometry} provides an analysis on UWB fused inertial based localization for aerial platforms, while Section~\ref{Visual inspection} establishes the 3D reconstruction problem from multiple images and multiple MAVs. Section~\ref{Experimental} demonstrates the experimental setup and presents the experimental trials for the presented inspection system. Finally, the concluding remarks are presented in Section~\ref{Conclusions}.
%
\section{Aerial Inspection System}\label{Aerial Inspection System}
\begin{figure}[htbp]
\centering
\includegraphics[width = 0.5\linewidth]{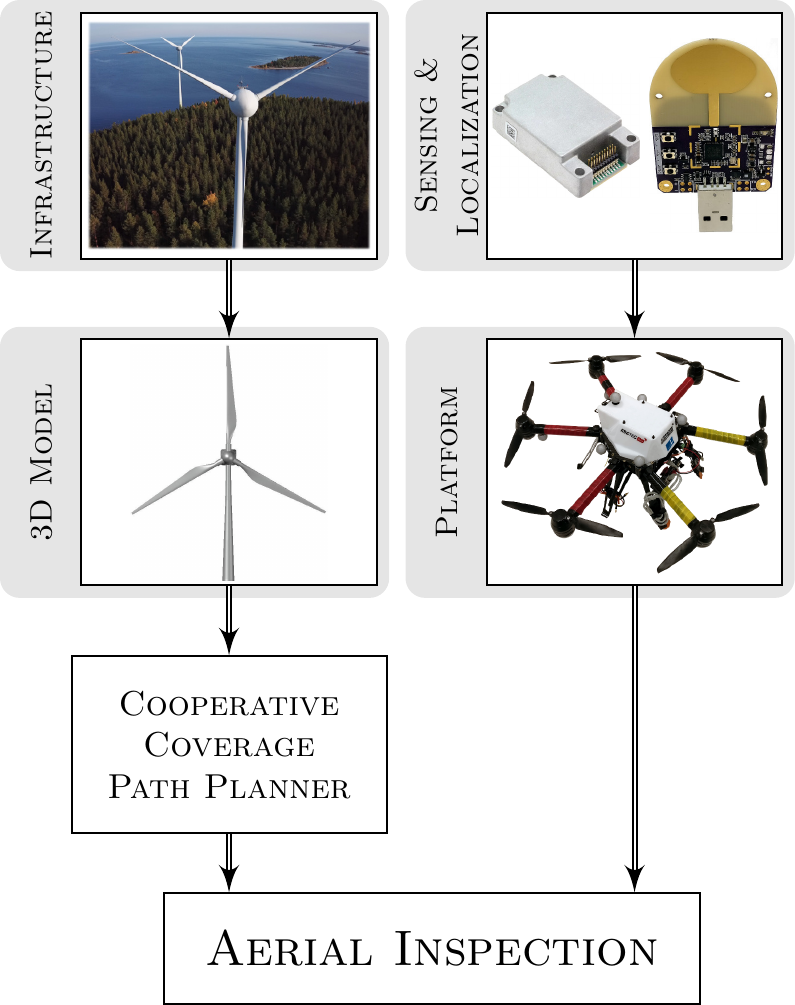}
\caption{An overview of the overall inspection system}
\label{fig:overallscheme}
\end{figure}

This article, inspired by the increasing capabilities of MAVs, establishes an autonomous aerial inspection system, which is specialized in large scale industrial facilities. The system is realized by either a single agent or a team of agents and is characterized by advanced localization and structure coverage capabilities, all demonstrated in real life by inspecting a wind turbine power plant, where the aim of the system is to provide visual data to infrastructure owners for further analysis and asset management. The overall scheme of the proposed system is depicted in Figure~\ref{fig:overallscheme}. 

\subsection{Field Trials and Open Challenges}\label{Challenges}
During the development of the proposed aerial inspection framework, the wind turbine site located in Bure\r{a}, Sweden have been visited multiple times. In these sites, the wind speed was measured up to 13~$\unit{m/s}$, while the wind turbine structure depicted in Figure~\ref{inspection_concept}, had a base diameter of approximately 4.5~$\unit{m}$, with a top diameter of approximately 1.5~$\unit{m}$ and with the height of the tower being 64~$\unit{m}$. Moreover, the length of each blade was 22~$\unit{m}$ with a corresponding cord length, at the root of the blade, of approximately 2~$\unit{m}$ and at the top of the blade of 0.2~$\unit{m}$, while the length of the hub and nacelle was approximately 4~$\unit{m}$.

\begin{figure}[htbp]
    \centering
        \includegraphics[width=0.5\linewidth]{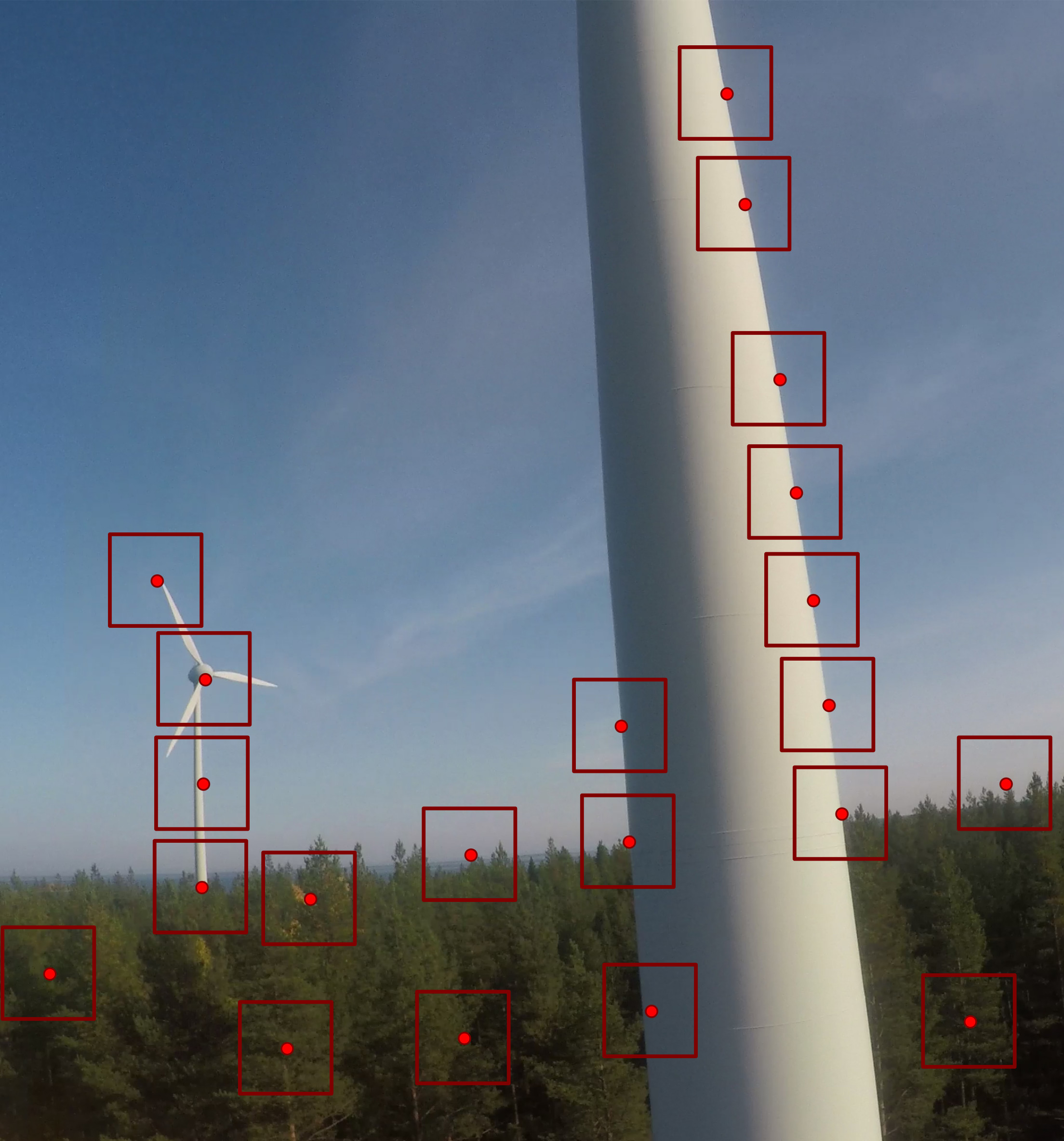}
        \caption{A sample of features detected on the wind turbine tower and the surrounding environment.}
        \label{fig:VIfeature}
    \end{figure}

Operating MAVs outside the lab, and especially around large scale infrastructures such as wind turbines, raises significant multidisciplinary research issues where one of the most important is to provide an accurate localization system that at the same time would be easily deployable. At the wind turbines the GPS solution fails at low height due to the multipath errors, which happen when the GPS receiver cannot distinguish a direct signal from a reflection, a fact that causes significant errors in the measurements. Usually, the GPS works well in positions where the interference from the building is small enough, however this is only at significant heights in this case. Moreover, the trending technology of visual inertial odometry, opposite to GPS, cannot provide reliable localization feedback in high altitudes. These algorithms base part of their processing in visual measurements by detecting areas of high contrast and texture. More specifically, in high altitudes this processing becomes unreliable, since they cannot detect and extract distinctive features from the environment due to lack of feature-rich local surfaces/areas, e.g. in the case of wind turbines which are simply described by a flat white color. This makes it difficult for the visual inertial odometry software to converge its state of movement to the actual state. As depicted in Figure~\ref{fig:VIfeature}, the detected features are far-away, while there are no features on the wind turbine tower itself except for unstable boundary features, and egomotion causes very little feature movement to the background.

Furthermore, the challenges of the visual sensors, identified for localization, extends also to other visual processing tasks, such as 3D reconstruction, where during the performed experimental trials it was found that the depth and stereo sensors failed to provide a solid 3D model of the wind turbine. Additionally, MAVs provide a limited flight time, which can be affected by external disturbances, such as wind gusts, payloads and temperature of the environment. This limits the feasibility of the inspection mission with one MAV, especially in large scale structures, such as wind turbines. Moreover, strong wind gusts cause significant drift of the MAV from the predefined trajectory and it should be compensated by the MAV's position controller. Thus the limited flight time and the deviation from the trajectory should taken under consideration or the overall system can fail to perform the inspection, or even worse result in a collision with the infrastructure that might cause damages to both the infrastructure and the aerial platform.

\subsection{System Hardware}
\subsubsection{MAV}
\begin{figure}[!htbp]
    \centering
    \begin{tikzpicture}
      	\node at (0,0) [] {\includegraphics[width=0.7\columnwidth]{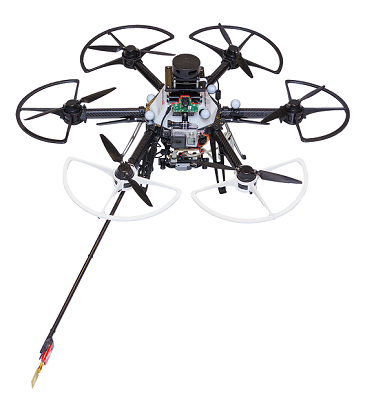}};

	    \node at (-1.4,-4.3) [right] {UWB node};
        \draw[-latex,line width = 0.03cm,white!50!black,dashed] (-1.5,-4.3) -- (-3.1,-3.8);
        
        \node at (0,-1.3) [] {VI sensor};
        \draw[-latex,line width = 0.03cm,white!50!black,dashed] (0,-1) -- (0,0.7);
        
        \node at (3,1.2) [right] {GoPro};
        \draw[-latex,line width = 0.03cm,white!50!black,dashed] (2.9,1.2) -- (0.3,1.5);
        
        \node at (3,4.5) [] {PlayStation camera};
        \draw[-latex,line width = 0.03cm,white!50!black,dashed] (3,4.3) -- (0.3,2.4);
        
        \node at (-3,4.5) [] {LIDAR};
        \draw[-latex,line width = 0.03cm,white!50!black,dashed] (-3,4.3) -- (-0.3,3.3);
        
        \draw[latex-latex,line width = 0.03cm,white!50!black,dashed] (-3.8,-0.6) -- (3.9,-0.6);
        \node at (2.5,-0.85) [] {0.87m};
    \end{tikzpicture}
    \caption{AscTec NEO platform equipped with the utilized full sensory system for the aerial inspection.}
    \label{fig:NEOcombined2}
\end{figure}
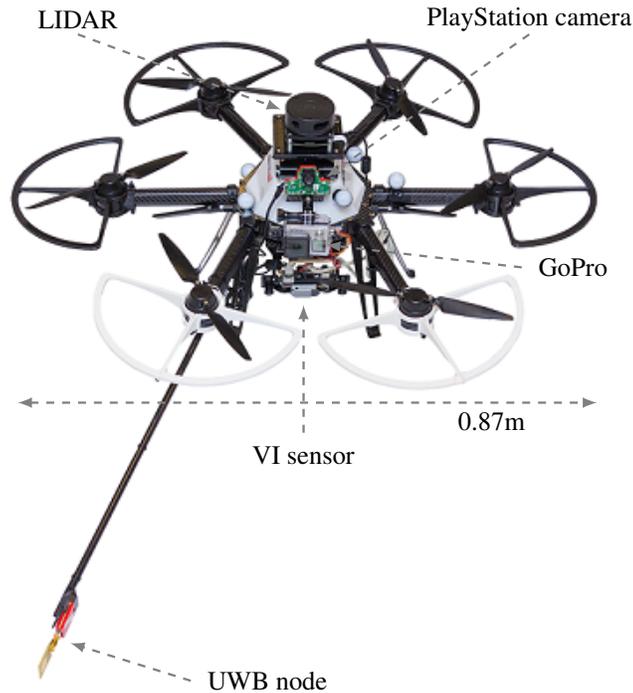

For the envisioned aerial inspection system for large scale infrastructures, the Ascending Technologies NEO hexacopter was utilized as the MAV platform, where in Figure~\ref{fig:NEOcombined2} the overall specifications and the selected sensors are presented. This platform is capable of providing a flight time of up to 26\,$\unit{min}$ without payload and in ideal conditions, with a maximum payload capacity up to 2\,$\unit{kg}$. For onboard processing, the belly of the MAV contains an Intel~NUC computer with a Core i7-5557U and 8 GB of RAM that runs Ubuntu Server 16.04 with the Robotic Operatic System (ROS) as its core. The platform has been equipped with a large set of different sensors, as depicted in Figure~\ref{fig:NEOcombined2}, where each component will be explained in the sequel.
    
\subsubsection{Localization system}
Due to the feature-less surface of the wind turbines for visual odometry and the existence of multipath errors in the GPS measurements, as was discussed in the prequel, the localization algorithms based on cameras and GPS failed during the field trials, and thus the proposed localization system was based on UWB and IMU fusion. This component is extensively explained in Section~\ref{Inertial Odometry}.

\subsubsection{Sensor suite}
The proposed sensory suite for the aerial inspectors included 3 different cameras: a) the Visual-Inertial (VI) sensor, b) the GoPro Hero4, c) the PlayStation Eye, and an additional laser range finder RPLIDAR, as depicted in Figure~\ref{fig:NEOcombined2}. The VI sensor developed by Skybotix AG with a weight of 0.117\,$\unit{kg}$ was attached below the hexacopter with a 45\,$\unit{^\circ}$ tilt from the horizontal plane, which is a monochrome, high dynamic range, global shutter stereo camera with 120\,$\unit{^\circ}$ DFOV and with a resolution of 752x480 pixels, moreover it is housing an Analog Devices ADIS16445 tactical grade IMU. Both cameras and IMU were tightly temporally aligned with hardware synchronization, while the cameras were operated at 20\,$\unit{fps}$. The GoPro Hero4 camera was attached on top of the hexacopter facing forward with a weight of 0.2\,$\unit{kg}$, while it was capable of recording high-definition video at various resolutions, ranging from 720p to 4000p and at a rate of 15-120\,$\unit{fps}$, while during the experimental trials the camera was operated with a 2K resolution at 30\,$\unit{fps}$. The Playstation Eye camera was attached in the middle of the hexacopters housing, facing forward with a weight of 0.150\,$\unit{kg}$, this camera was operated at 20\,$\unit{fps}$ and with a resolution of 640x480 pixels. The variety in the specifications of the camera suite was motivated by the need to test their performance under challenging conditions, regarding the dataset collection. Thus, the main aim was to use the captured frames for direct visual inspection by experts in the structure maintenance, while the data from the VI sensor and the GoPro camera were also used to provide 3D models of the inspected parts. Finally, RPLDAR was a low cost laser sensor, which provides a $360^\circ$ scan field at a 5.5\,$\unit{Hz}$/10\,$\unit{Hz}$ rotating frequency with guaranteed 8 meter range. This laser scanner has also been tested during the experimental trials for enabling the online obstacle avoidance schemes.

\subsection{Cooperative Coverage Path Planner}\label{Cooperative Coverage Path Planner}
%
Towards the vision of the inspector MAV, the theoretical framework established in~\cite{mansouri2018cooperative} is integrated in the autonomous inspector framework and experimentally tested in the complex case of a windturbine structure. Briefly, the coverage scheme is capable of providing a path for accomplishing a full coverage of the infrastructure, without any shape simplification, by slicing it by horizontal planes to identify branches of the infrastructure and assign specific areas to each agent. Complicated structures have multiple branches e.g. in wind turbine the base and each blade are considered as branches, where the proposed method identifies these branches and assign paths to $n$ agents. If the structure has one branch all $n$ agents are assigned to the same branch, otherwise the $n$ agents are equally distributed to different branches. Furthermore, to guarantee a full coverage to facilitate visual processing, the introduced path planning creates for each agent an overlapping visual inspection area. The novel established C-CPP scheme, in addition to the position references, provides also yaw references for each agent to assure a field of view, directed towards the structure surface. 

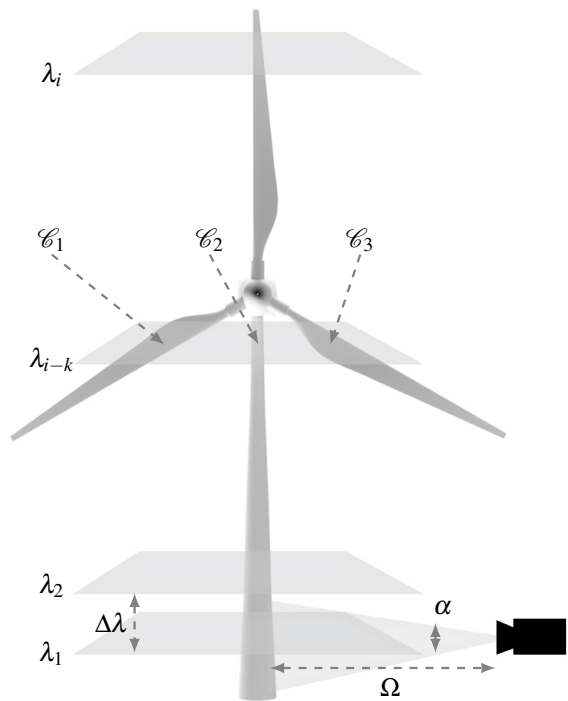
\begin{figure}[!htbp]
  \centering
  \begin{tikzpicture}
    \node at (0,0) [] {\includegraphics[width = 0.66\linewidth]{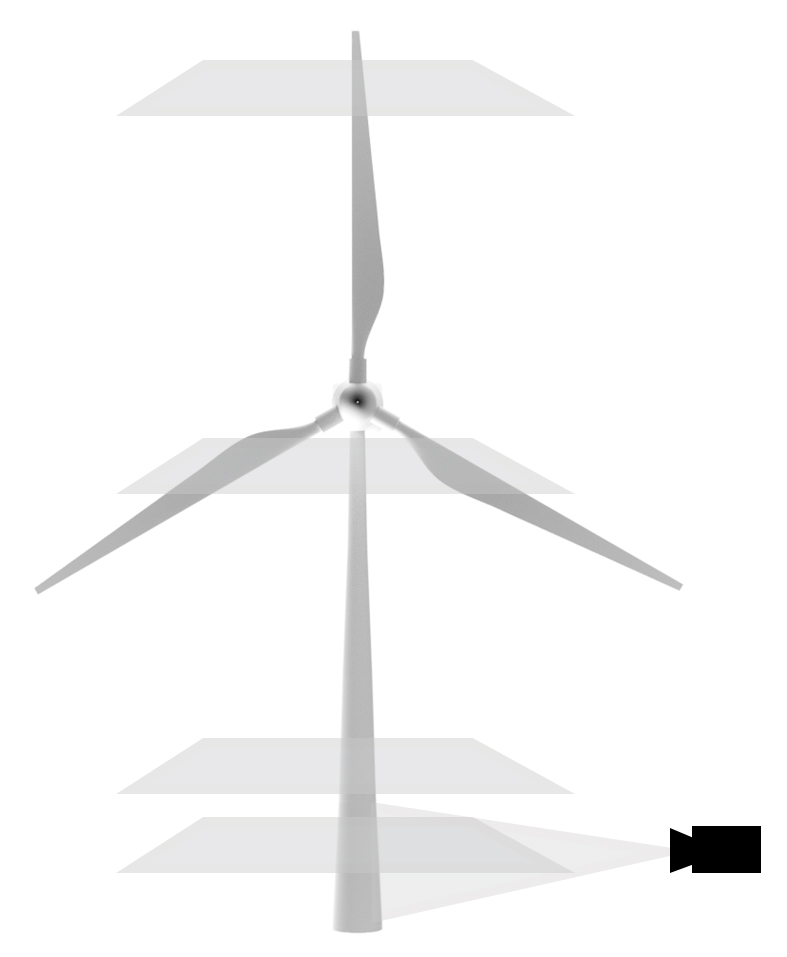}};
    
  \node at (-3.1,-4) [] {$\lambda_1$};
  \node at (-3.1,-3.1) [] {$\lambda_2$};
  \node at (-3.1,-0.1) [] {$\lambda_{i-k}$};
  \node at (-3.1,3.7) [] {$\lambda_i$};
  
  \node at (-3.1,1.5) [] {$\mathcal{C}_1$};
  \draw[-latex,line width = 0.03cm,black!50!white,dashed] (-3.1,1.3) -- (-1.6,0.1);
  \node at (-1,1.5) [] {$\mathcal{C}_2$};
  \draw[-latex,line width = 0.03cm,black!50!white,dashed] (-1,1.3) -- (-0.35,0.1);
  \node at (1,1.5) [] {$\mathcal{C}_3$};
  \draw[-latex,line width = 0.03cm,black!50!white,dashed] (1,1.3) -- (0.6,0.1);
  
  \node at (1.4,-4.45) [] {$\Omega$};
  \draw[latex-latex,line width = 0.03cm,black!50!white,dashed] (-0.2,-4.2) -- (2.8,-4.2);
  
  \node at (2.1,-3.4) [] {$\alpha$};
  \draw[latex-latex,line width = 0.03cm,black!50!white,dashed] (2,-3.6) -- (2,-4);
  
  \node at (-2.3,-3.6) [] {$\Delta \lambda$};
  \draw[latex-latex,line width = 0.03cm,black!50!white,dashed] (-2,-3.2) -- (-2,-4);
  
  \end{tikzpicture}
  \caption{An overview of the mathematical notations used in the C-CCP algorithm.}
  \label{fig:cccp_defs}
\end{figure}

For the use of the C-CPP, initially the general case of an aerial inspector equipped with a limited Field of View (FOV) sensor was considered, determined by an aperture angle $\alpha$ and a maximum range $r_{max}$. Furthermore, $\Omega \in \mathbb{R}^+$ is the user-defined offset distance ($\Omega < r_{max}$), from the infrastructure's target surface and $\Delta \lambda$ is the distance between each inspected plane. $\Delta \lambda$ is equal to $\frac{\Omega}{\beta} \tan{\alpha/2}$, where the parameter $\beta \in [1,+\infty)$ represents the ratio of overlapping. The horizontal planes are defined as $\lambda_i$, with $i\in \mathbb{N}$. The 3D map of the infrastructure is provided as a set $\boldsymbol{S}$ with a finite collection of points, denoted as $\boldsymbol{S}=\{p_i\}$, and $p_i=[x_i,y_i,z_i]^\top \in \mathbb{R}^3$. Furthermore, $\mathcal{C}_j(x,y,z)$ with $j \in [1,m]$ are the points in each branch and $m$ is the overall number of branches in the structure. A graphical overview of this C-CCP scheme is presented in Figure~\ref{fig:ccp}.
\tikzstyle{decision} = [diamond, draw, 
    text width=8em, text badly centered, node distance=3cm, inner sep=0pt, minimum height=2em]
\tikzstyle{block2} = [rectangle, draw,  
    text width=10em, text centered, rounded corners, minimum height=4em]
\tikzstyle{block3} = [rectangle, draw,  
    text width=8em, text centered, rounded corners, minimum height=2em]
    \tikzstyle{block} = [rectangle, draw, 
    text width=8em, text centered, rounded corners, minimum height=4em]
\tikzstyle{line} = [draw, -latex']
\tikzstyle{cloud} = [draw, ellipse, node distance=3cm,
    minimum height=2em]
\begin{figure}[!htbp] \centering
\resizebox{0.4\linewidth}{!}{%
\begin{tikzpicture}[node distance = 2cm, auto]
    \node [block2] (init) {require parameters: $\boldsymbol{S}(x,y,z)$, $\Delta \lambda$,  $\Omega$, $n$, $d_s$, $T_s$, $\vec{V}_d$}; 
    \node [cloud, left of=init] (User) {User};
    \node [block, below of=init] (identify) {slice the structure by horizontal planes $\lambda_i$};
    \node [block, below of=identify] (clusters) {identify the branches $\mathcal{C}(x,y,z)$};
    \node [block, below of=clusters] (offset) {add safety distance $\Omega$};
    \node [block, below of=offset] (pathassign) {path assignment to the $n$ agent};
    \node [block3, left of=clusters, node distance=4cm] (update) {$\lambda_{i+1}= \lambda_{i}+ \Delta \lambda$};
    \node [decision, below of=pathassign] (decide) {$\lambda_i \le max_z \boldsymbol{S}(x,y,z)$};
    \node [block3, below of=decide, node distance=3cm] (stop) {STOP};
    \path [line] (init) -- (identify);
    \path [line] (identify) -- (clusters);
    \path [line] (clusters) -- (offset);
    \path [line] (offset) -- (pathassign);
    \path [line] (pathassign) -- (decide);
    \path [line] (decide) -| node [near start] {yes} (update);
    \path [line] (update) |- (identify);
    \path [line] (decide) -- node {no}(stop);
    \path [line,dashed] (User) -- (init);
\end{tikzpicture}
}
\caption{Flowchart of the overall C-CPP scheme.}
\label{fig:ccp}
\end{figure}
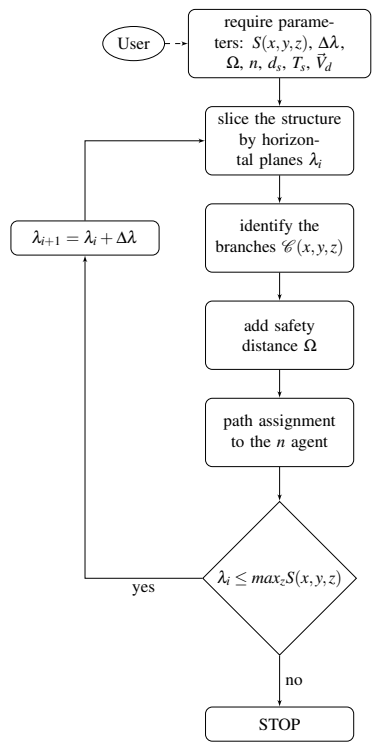

\subsection{UWB Inertial Odometry Framework} \label{Inertial Odometry}
UWB Radio Frequency (RF) communication is based on using a wide band of the RF spectrum, rather than a single frequency as a carrier wave radio does, which has the temporal representation of a pulse and as a result is sometimes referred to as a pulse radio. Due to the high center frequency (3.1~to~4.8~GHz and 6.0~to~10.6~GHz) and the spectral width of the pulse (499.2~to~1331.2~MHz) the pulses have good spatial resolution, which makes them ideal for time stamping RF packets, referred to as messages, with high accuracy. This property of accurate timestamps, together with good reference clocks, give the ability to estimate the distance between two transceivers by exchanging 2 or more packets and thus it could be considered that the distance estimation is a byproduct of communication.

Furthermore, one major drawback of a carrier wave based radio is the problem of multipathing, where the carrier wave forms destructive interference with itself, effectively reducing the received signal strength, or introducing an unknown phase shift. This is a problem that is severely mitigated in the UWB radio, where the spatial length of each pulse is small enough for each pulse to be detected uniquely and this allows the receiver to reconstruct the pulse from multiple reflections. In a sense, the more the reflections are available, the stronger the received signal is, in contrary to GPS, which can give highly misleading measurements when close to tall structures. 

In Figure~\ref{fig:NEOcombined2}, the UWB node developed by LTU is depicted when mounted on the MAV. This hardware contains all the embedded electronics including the microprocessor, 3-axis accelerometer, 3-axis gyroscope, the UWB RF transceiver and the antenna to enable the UWB communication and localization, while this system is fully self contained and can directly be deployed for enabling full localization of the MAV state.

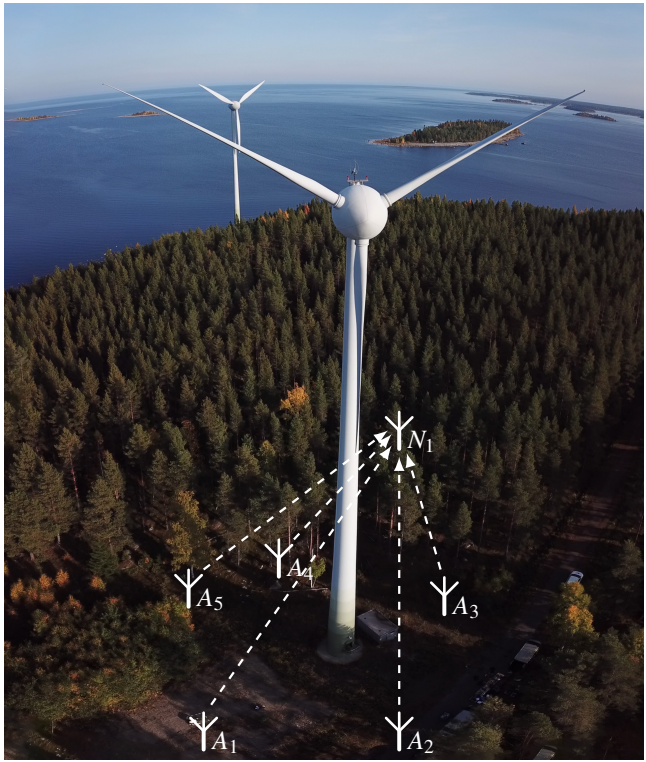
\begin{figure}[!htbp]
    \centering
    \begin{tikzpicture}
      \node at (0,0) [] {\includegraphics[width=0.7\columnwidth]{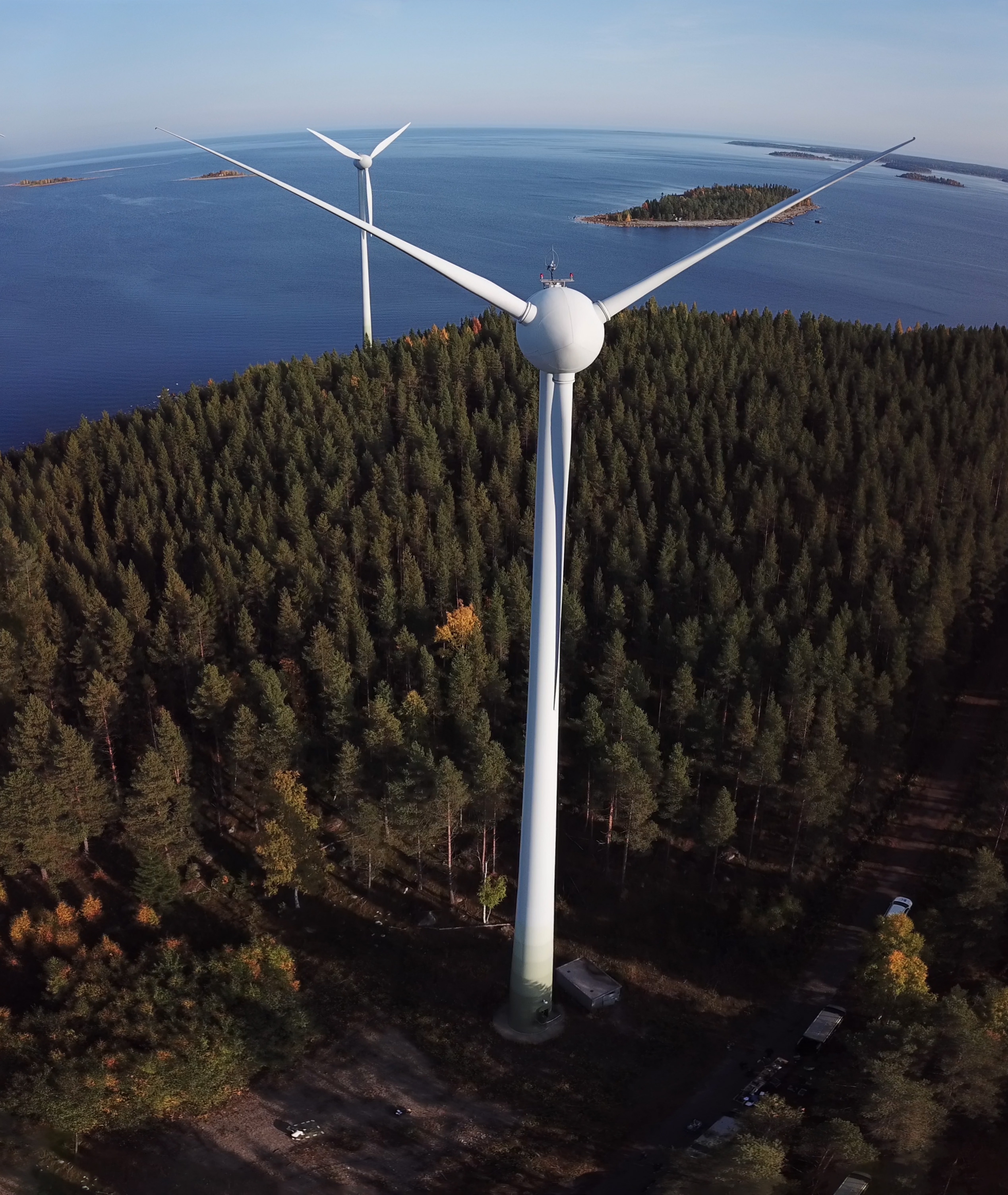}};

	    \antenna{(1.0,-1.0)}{0}{3.5}{white} 
        \node at (1.3,-0.8) {\textcolor{white}{$N_1$}};
        
		\antenna{(-1.6,-5.0)}{0}{3.5}{white} 
        \node at (-1.3,-4.8) {\textcolor{white}{$A_1$}};
        \draw[-latex,line width = 0.03cm,white,dashed] (-1.5,-4.3) -- (0.9,-0.85);
        
        \antenna{(1.0,-5.0)}{0}{3.5}{white} 
        \node at (1.3,-4.8) {\textcolor{white}{$A_2$}};
        \draw[-latex,line width = 0.03cm,white,dashed] (1.0,-4.3) -- (1,-1.0);
        
        \antenna{(1.6,-3.2)}{0}{3.5}{white} 
        \node at (1.9,-3.0) {\textcolor{white}{$A_3$}};
        \draw[-latex,line width = 0.03cm,white,dashed] (1.55,-2.5) -- (1.1,-0.95);
        
        \antenna{(-0.6,-2.7)}{0}{3.5}{white} 
        \node at (-0.3,-2.5) {\textcolor{white}{$A_4$}};
        \draw[-latex,line width = 0.03cm,white,dashed] (-0.4,-2.1) -- (0.9,-0.7);
        
        \antenna{(-1.8,-3.1)}{0}{3.5}{white} 
        \node at (-1.5,-2.9) {\textcolor{white}{$A_5$}};
        \draw[-latex,line width = 0.03cm,white,dashed] (-1.6,-2.5) -- (0.85,-0.65);
    \end{tikzpicture}
    \caption{An overview of the UWB localization system, where $A_1$ - $A_5$ are the stationary anchors and $N_1$ is the tracked node mounted on the MAV, while the dashed lines highlight the measured distances.}
    \label{fig:uwbsys}
\end{figure}

For a proper operation of the estimation framework, it is needed to have the UWB transceivers with known and fixed positions, called anchors (while the transceiver on the MAV is called a tag), spread out in the working area to act as known positions to measure distances for the later trilateration and fusion with the IMU, as described in \cite{fresk2017uwb}. This is directly analogous to GPS, while here the “satellites” (anchors) are placed as needed within the operating volume, conceptually presented in Figure~\ref{fig:uwbsys}.

\subsection{Surface reconstruction} \label{Visual inspection}
As stated in the prequel, this work targets the application scenario of autonomous inspection by single or multiple MAVs, where the objective of the inspection missions is the collection of high resolution visual data of regions of interest and the generation of 3D surface models. All available data will be used afterwards by inspection experts to analyze and detect possible defects on their assets. To this end, each aerial platform is equipped, but not limited, with a camera to record the required data from the infrastructure. 

During the navigation of the MAVs around the structure, the raw visual stream is directly available for defect assessment. Regarding the surface reconstruction, the main approach to process the data considers a monocular camera Structure from Motion (SfM)~\cite{schonberger2016structure}, where the MAVs fly around covering specific parts, with the aim to collaboratively process all the captured data into a global representation. The selection of monocular mapping is driven by the application scale and the object characteristics. Generally, the perception of depth using stereo cameras is bounded to the stereo baseline, essentially reducing the configuration to monocular at far ranges and to this end, stereo algorithms cannot perform in cases with large structures and high altitudes. The employed SfM approach is an offline process that provides a sparse 3D reconstruction and accurate MAV poses, by using different camera viewpoints and consists of a massive optimization process. Finally, the data collected during the navigation mission is down sampled, since they contain redundant information from all the camera frames and there is a need to keep the resulting outcome within a reasonable time, while the sparse pointcloud is inserted into Multi View Stereo algorithms to perform dense reconstruction based on multi view stereo pairs~\cite{seitz2006comparison}.
 
\subsection{System Software}

\begin{figure*}[!htbp]
\centering
\includegraphics[width = 0.9\linewidth]{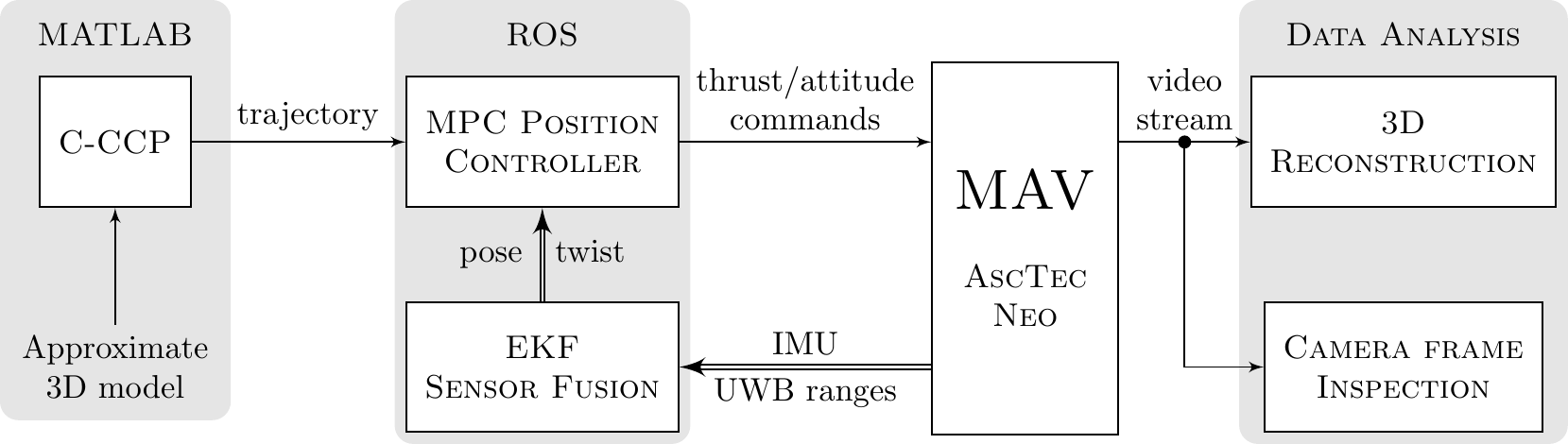}
\caption{An overview of the proposed aerial inspection system's software architecture.}
\label{fig:experimental_setup}
\end{figure*}

The navigation system of the aerial inspector is integrated within the ROS framework, where two main components provide autonomous flight, namely an UWB inertial odometry estimator, based on the Multi Sensor Fusion Extended Kalman Filter (MSF-EKF)~\cite{msf} and a linear Model Predictive Control (MPC) based position controller~\cite{mav_linear_mpc_}. The sensor fusion node consists of an EKF filter that does tight inertial fusion from the hexacopter's IMU during the state propagation and the UWB range measurements are utilized during the filter correction step. The outcome of the UWB inertial odometry are the position, orientation (pose), the linear/angular velocity (twist) of the aerial robot and the IMU biases. This consists of an error state Kalman filter performing sensor fusion as a generic software package that has the unique feature to handle delayed measurements, while staying within the desired computational bounds. The linear MPC position controller~\cite{mav_linear_mpc_} generates attitude and thrust references for the NEO's predefined low level attitude controller, with the aim to have separation of concerns, as the high level control and planning algorithms should have minimal knowledge of the low level controllers. The overall functional schematic of the experimental setup is presented in Figure \ref{fig:experimental_setup}.

The C-CPP method, described in Section \ref{Cooperative Coverage Path Planner}, has been entirely implemented in MATLAB. The inputs for the method are a 3D model of the infrastructure of interest and specific parameters, which are the number of agents ($n$), the offset distance from the object ($\Omega$), the FOV of the camera ($\alpha$), the desired velocity of the aerial robot ($V_d$) and the position controller sampling time ($T_s$). The generated paths are sent to the NEO platforms through the utilization of the ROS framework. 

\section{Experimental results}\label{Experimental}

\subsection{Mission Preliminaries}

The presented aerial platform with the sensor systems and combined with the developed algorithmic components, described in previous section, constitutes the autonomous aerial inspection system named autonomous aerial inspector. The capabilities of the aerial inspectors have been publicly demonstrated for the case of wind turbine inspection in Sweden, where the mission scenario was two-fold by targeting the inspection of two separate parts of the structure, namely the wind turbine tower and the wind turbine blades. The requirements for the system were to provide a complete coverage of the inspected parts autonomously, while storing all necessary visual data for further analysis. Although, two agents were used for the specific case presented in this work, the presented inspection system can operate either in a single agent or multi-agent mode, depending on the application needs and the flying limitations of the MAVs.   

The initial step for the deployment of the inspection system was to setup the ground station for monitoring the operations and fix 5 UWB anchors around the structure, with specific coordinates presented in Table~\ref{tab:UWB anchor coordinates}, which constitute the infrastructure needed for the localization system of each aerial platform. The number of anchors as well as their position has been selected in a manner to guarantee UWB coverage around all parts of the wind turbine. From a theoretical point of view~\cite{fresk2017uwb}, only 3 anchors are needed, however it is common that one anchor will be behind the wind turbine for the MAV's point of view, which gives rise to a minimum of 4 anchors to compensate, while a fifth anchor was added as redundancy. The resulting fixed anchor positions provide a local coordinate frame that guarantees repeatability of the system, and with the significant ability to revisit the same point multiple times, in case the data analysis shows issues that require further inspection. An important note for all the inspection cases on the wind turbine and for the system in operation is that the blades are locked in a star position, as shown in Figure~\ref{fig:uwbsys}, which simplifies the 3D approximate modeling of the structure.

\begin{table}[!htbp]
\caption{UWB anchor placement locations.}
\begin{center}
      \begin{tabular}
        {
        >{\centering \arraybackslash} p{0.15\columnwidth} <{}
        >{\centering \arraybackslash} p{0.1\columnwidth} <{}
        >{\centering \arraybackslash} p{0.1\columnwidth} <{}
        >{\centering \arraybackslash} p{0.1\columnwidth} <{}
        >{\centering \arraybackslash} p{0.1\columnwidth} <{}
        >{\centering \arraybackslash} p{0.1\columnwidth} <{}
        }
        \textbf{Coordinate} & $\boldsymbol{A}_1$ & $\boldsymbol{A}_2$ & $\boldsymbol{A}_3$ & $\boldsymbol{A}_4$ & $\boldsymbol{A}_5$ \\ \hline
        $x$       & \unit[0]{m} & \unit[26.1]{m} & \unit[6.6]{m}  & \unit[19.5]{m} & \unit[14.6]{m}  \\
 		$y$       & \unit[0]{m} & \unit[0]{m}    & \unit[24.8]{m} & \unit[18.2]{m} & \unit[-21.7]{m} \\
 		$z$       & \unit[0]{m} & \unit[0]{m}    & \unit[0]{m}   & \unit[0]{m}    & \unit[0]{m}     \\ \hline
    \end{tabular}
\end{center}
\label{tab:UWB anchor coordinates}
\end{table}

In the proposed architecture, all the processing necessary for the navigation of the MAVs is performed onboard, while the overview of the mission and the commands from the mission operators (inspectors) is performed over a WiFi link, while the selection of WiFi is not a requirement and can be replaced with the communication link of choice e.g. 4G cellular communication. The UWB based inertial state estimation runs at the rate of the IMU, which in this case was 100 Hz, and the generated coverage trajectory has been uploaded to the MAV before take-off, which is followed as soon as the mission started by the command of the operator. The paths have been followed autonomously, without any intervention from the operators on the site, and the collected data have been saved onboard, while after downloading the mission data post processing is performed in the ground station or in the cloud. The data provided by the system can be used for position aware visual analysis, examining high resolution frames or they can be post-processed to generate 3D reconstructed models. The key feature to be highlighted from the inspection is that any detected fault can be fully linked with specific coordinates, which can be utilized by another round of inspections or for guiding the repair technician. The final, is a major contribution of the presented aerial inspection system, since this need is the fundamental information that is needed for enabling a safe and autonomous aerial inspection that has the potential to performed the human based ones.

\subsection{Wind turbine inspection}
For the specific case of wind turbines the C-CCP generated inspection paths have been obtained with two autonomous agents in order to reduce the needed flight time, and still be within the battery constrained flight time of the utilized MAV. However, due to the limited flight time of the MAVs in the field trials, the inspection problem has been split into the tower inspection and the blade inspection, where the specifics of each is presented in the sequel Table~\ref{tab:info_mission}, while both can be performed at the same time with more MAVs to reduce inspection time even further. A common characteristic for both of the cases is that the generated path for each MAV keeps a constant safety distance from the structure, while at the same time is keeping it in view of the visual sensors, and maximizing the safety distance between agents, which gives rise to the agents being on opposite side of the wind turbine at all times. The area in which the inspection is performed is generally of high wind and while the inspection of the tower is protected from wind, owing to the forest, the blade inspection is above the tree line. Thus, the aerial inspections have been specifically tuned to compensate strong wind gusts that were measured up to 13 m/s, where the tunning was targeting the MAV's controller’s weight on angular rate that has been increased to significantly reduce the excessive angular movement.

\begin{table}[!htbp]
\caption{Overview of the inspection configurations.}
\begin{center}
    \begin{tabular}
        {
        >{\centering \arraybackslash} p{0.45\columnwidth} <{}
        >{\centering \arraybackslash} p{0.15\columnwidth} <{}
        >{\centering \arraybackslash} p{0.15\columnwidth} <{}
        }
        \textbf{Mission Configuration} &  \textbf{Tower} & \textbf{Blade} \\ \hline
        Number of agents      & 2         & 1         \\
        Inspection Time       & \unit[144]{sec}	  & \unit[206]{sec} \\
        Safety Distance       & \unit[7]{m} 		  & \unit[9]{m} \\
        Velocity              & \unit[1]{m/s} 		  & \unit[1.2]{m/s}  \\
        Starting height       & \unit[8]{m}  		  &\unit[30]{m}   \\
        Finishing height      &  \unit[24]{m} 		  &\unit[45]{m}   \\ \hline
    \end{tabular}
\end{center}
\label{tab:info_mission}
\end{table}

\subsubsection{Tower inspection}
\begin{figure}[!htbp]
  \centering
  \begin{tikzpicture}
    \node at (0,0) [align=center] {\includegraphics[width = 0.74\linewidth]{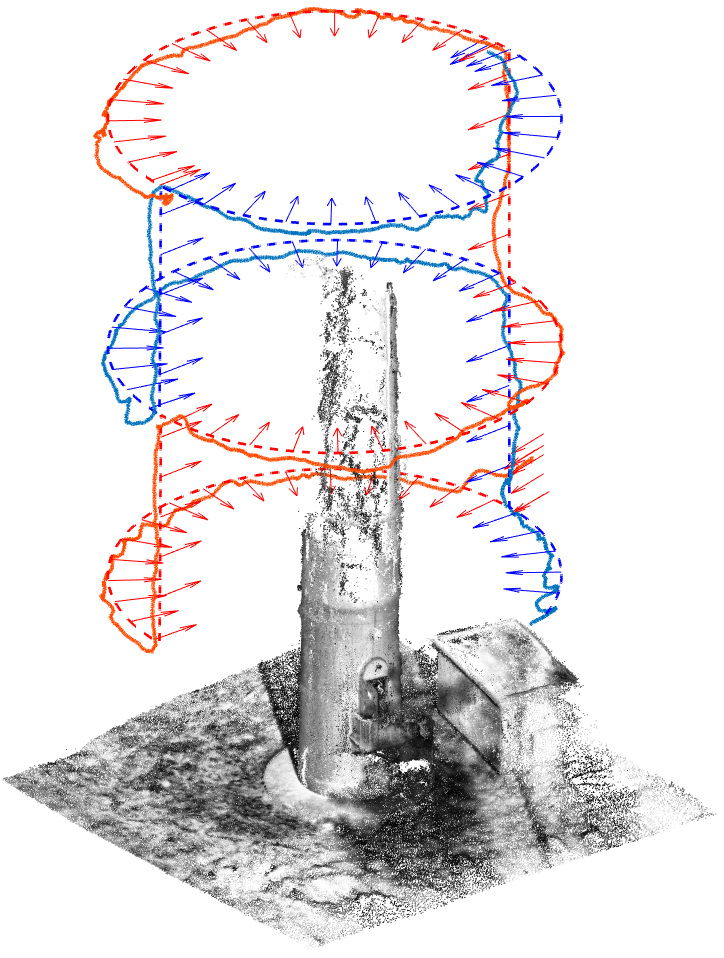}\\ \\ \includegraphics[width = 0.31\linewidth, cfbox=black 0.2mm -0.1mm]{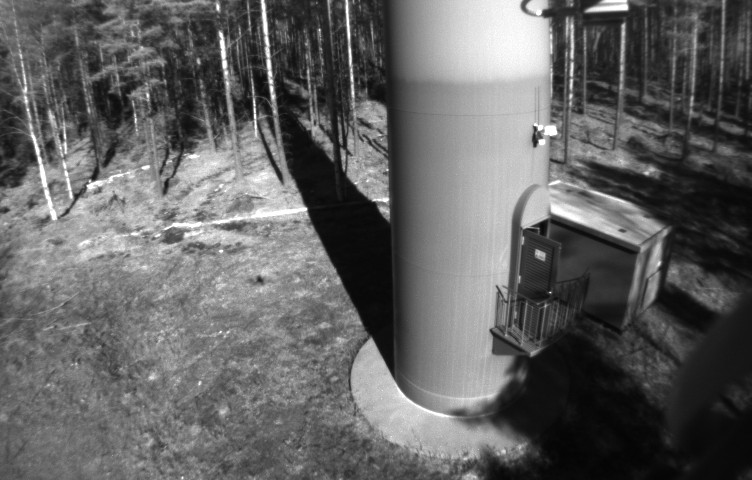} \includegraphics[width = 0.31\linewidth, cfbox=black 0.2mm -0.1mm]{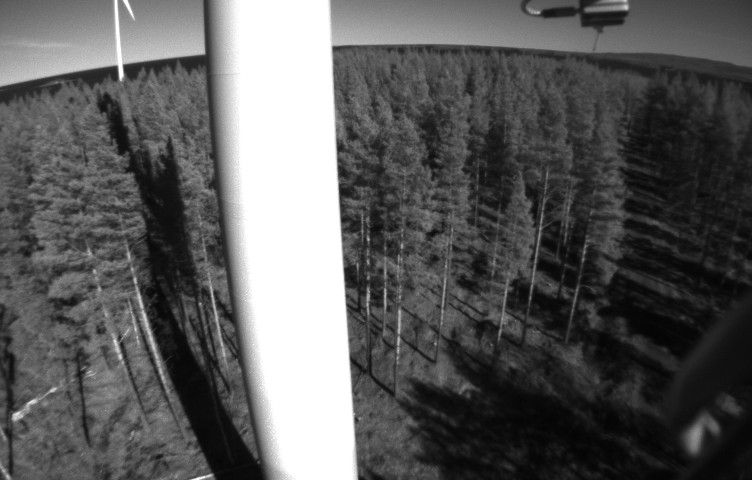} \includegraphics[width = 0.31\linewidth, cfbox=black 0.2mm -0.1mm]{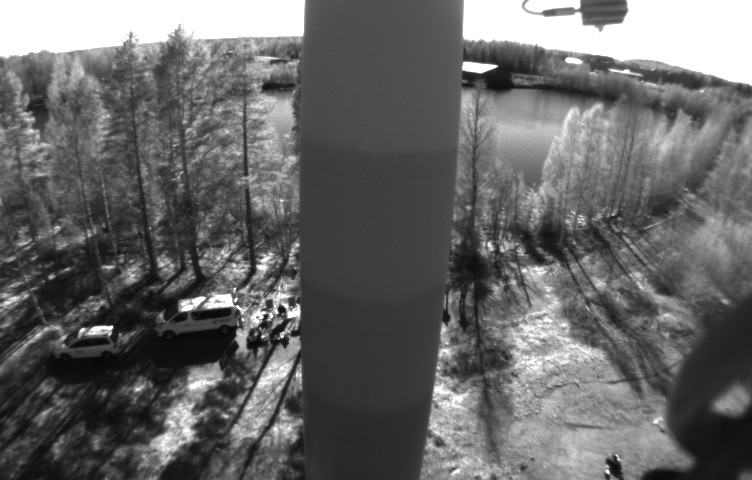}};
    
  \draw[-*,line width = 0.03cm,white!30!black,dashed] (-3,-4.9) -- (-2.5, -0.3);
  \draw[-*,line width = 0.03cm,white!30!black,dashed] (0,-4.9) -- (-2.4,5.3);
  \draw[-*,line width = 0.03cm,white!30!black,dashed] (3,-4.9) -- (1.3,4.25);
  
  \end{tikzpicture}
  \caption{Coverage paths followed by 2 agents with actual (solid) and reference paths (dashed) together with desired direction, which resulted in the depicted 3D reconstruction and sample camera frames of the base and tower to be used by the inspector. }
  \label{fig:path_reconst_tower}
\end{figure}

In the specific case of the wind turbine base and tower inspection, the generated paths are of a circular shape, as depicted in Figure~\ref{fig:path_reconst_tower}, which is the result of the constant safety distance from the structure based on the C-CCP algorithm. As can be seen from the tracked trajectories the controllers perform well with an RMSE of 0.5464 m, while at the top of the trajectory a more significant error can be seen that is induced from the specific MAV transitioning above the tree-line, where a wind gust caused the deviation from the desired trajectory where the MAV compensates and finishes it’s inspection trajectory. 

From the depicted reconstruction in Figure~\ref{fig:path_reconst_tower}, it is possible to understand that the base of the wind turbine, which is feature rich, provides a good reconstruction result, while as the MAV continues to higher altitudes, the turbine tower loses texture due to its flat white color, causing the reconstruction algorithms to not provide a successful reconstruction. However, the visual camera streams do have position and orientation for every frame, as depicted in Figure~\ref{fig:path_reconst_tower} for some instances, which allows for a trained inspector to review the footage and be able to determine if there are spots which need extra inspection or repairs. For the reconstruction in Figure~\ref{fig:path_reconst_tower}, the~\cite{pizer1987adaptive} and~\cite{schonberger2016structure} algorithms have been used, the former for pre-processing the images for enhance their contrast, while the latter was the SfM approach for providing the 3D model of the structure. The reconstruction took place on a PC with the configuration i7-7700 CPU and 32 GB of RAM, where the processing lasted approximately 4 hours.

\subsubsection{Blade inspection}
\begin{figure}[!htbp]
  \centering
  \begin{tikzpicture}
    \node at (0,0) [align=center] {\includegraphics[width = 0.7\linewidth]{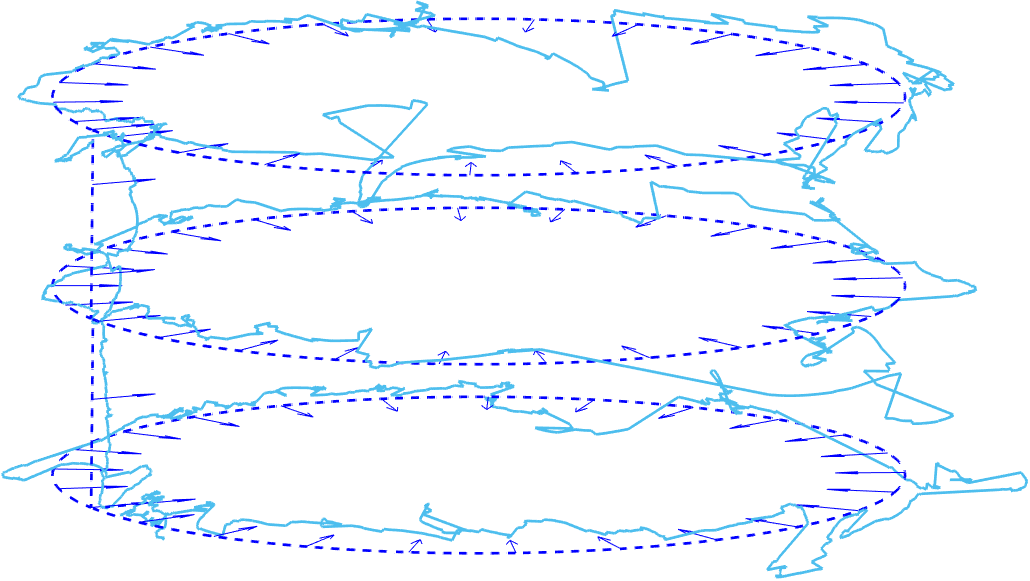}\\ \\ \includegraphics[width = 0.31\linewidth, cfbox=black 0.2mm -0.1mm]{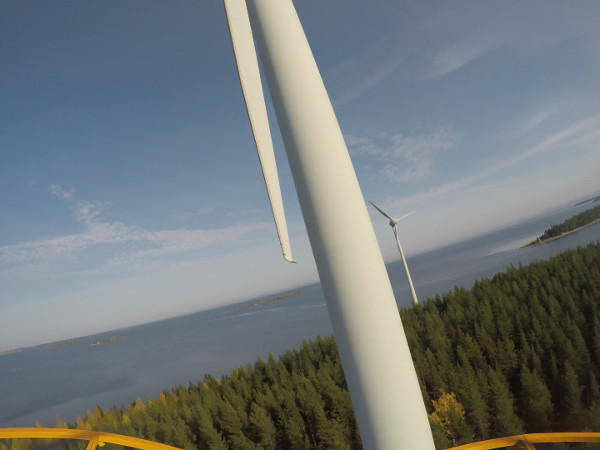} \includegraphics[width = 0.31\linewidth, cfbox=black 0.2mm -0.1mm]{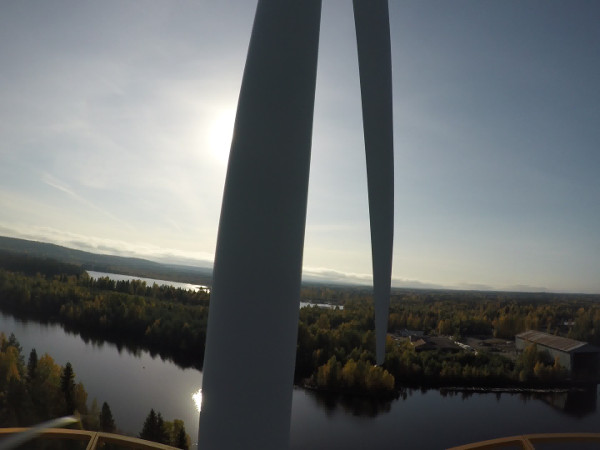} \includegraphics[width = 0.31\linewidth, cfbox=black 0.2mm -0.1mm]{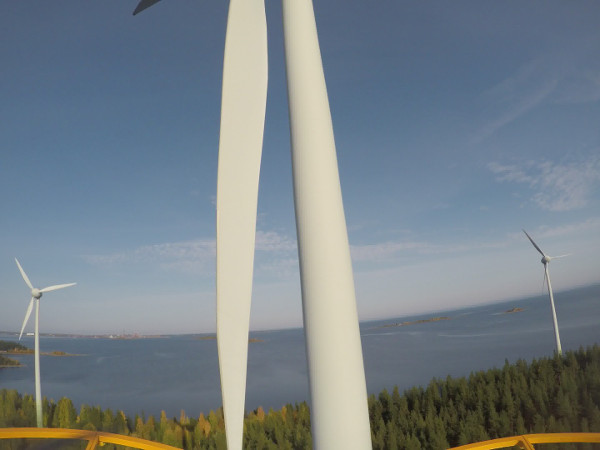}};
    
  \draw[-*,line width = 0.03cm,white!30!black,dashed] (-3,-1) -- (-2.5,-0.2);
  \draw[-*,line width = 0.03cm,white!30!black,dashed] (0,-1) -- (0.2,2.35);
  \draw[-*,line width = 0.03cm,white!30!black,dashed] (3,-1) -- (2.4,1.6);
  
  \end{tikzpicture}
  \caption{Coverage path followed by the agent with actual (solid) and reference path (dashed) together with desired direction, which resulted in the depicted 3D reconstruction and sample camera frames of the blade to be used by the inspector. Note the flat white color of the tower.}
  \label{fig:wind3D2}
\end{figure}

Compared to the base and tower inspection, for which the C-CCP algorithm generated circular trajectories, a similar approach was followed for the the blade inspections. This comes from the fact that the blade inspection is performed on the blade with a direction towards the ground and with the trailing edge of the blade towards the tower, which would cause the C-CCP algorithm to generate half-circle trajectories. However, in this case the same agent can inspect the final part of the tower by merging both tower and blade trajectories, as can be seen in Figure~\ref{fig:wind3D2}, while minimizing the needed flight time and demonstrating at a full extend the concept of aerial cooperative autonomous inspection. With the available flight time of the MAV, it is possible to inspect the blade with only one operating MAV, allowing for the safety distance between agents to be adhered to, by the separation of the inspected parts. However, during the blade inspection, the tracking performance of the MAV was reduced to an RMSE of 1.368 m, due to the constant exposure to wind gusts and the turbulences generated by the structure, and as these effects were not measurable, until the effects are observed on the MAV, it has reduced the overall observed tracking capabilities of the aerial inspectors. The second effect of the turbulence was the excessive rolling and pitching of the MAV, which introduced a significant motion blur in the captured video streams, due to the fixed mounting of the camera sensor, introducing the need for adding a gimbal for stabilizing the inspection camera and reducing the motion blur. Finally, as can be seen in the camera frames in Figure~\ref{fig:wind3D2}, there are no areas of high texture on the wind turbine tower or blades which caused 3D reconstruction to fail. However, the visual data captured is of high quality and suitable for review by an inspector.

\section{Lessons Learned}\label{lessons}
Throughout the experimental trials for this inspection scenario, many different experiences were gained that assisted in the development and tuning of the algorithms utilized. Based on this experience, an overview of the lessons learned is provided in the sequel with connections to the different utilized field algorithms.
\subsection{MAV Control}
When performing trajectory tracking and position control experiments indoors a dedicated laboratory many disturbances, which are significant in the field trials, can be neglected and this is especially true for strong wind gusts and turbulences caused by the structure. In the case of indoor experimental trials, the MAV can be tuned aggressively to minimize the position tracking error, while in the full scale outdoor experiments this kind of tuning would provide excessive rolling and pitching due to the controllers trying to fully compensate for the disturbances. However, this has the side effect of making the movements jerky and oscillatory, and overall reduce the operator’s trust in the system as it seems to be close to unstable. Furthermore, in the case that the controllers were tuned for a smooth trajectory following, larger tracking errors would have to be accepted in the trajectory following. During the field trials, some wind gust can even be above the operational limits of the MAV, causing excessive errors in the trajectory tracking. To reduce the effect in the outdoor experiments, the controller’s weight on angular rate was increased to significantly reduce the excessive movement, while in general the tuning of the high level control scheme, for the trajectory tracking, is a tedious task and it was found to be extremely sensitive to the existing weather conditions.
\subsection{Planning}
The path planner provides a path to guarantee for a full coverage of the structure, however in the field trials, due to high wind gusts, there are variations between the performed trajectory and the reference. Thus, there is a need for an online path planner for considering these drifts and re-plan the path or to have a system that it is able to detect if a specific part of the structure has been neglected and provides extra trajectories to compensate. Additionally, due to the payload, the wind gusts and the low ambient temperature, the flight time was significantly less than the expected value from the MAV manufacturer. In certain worst cases, this time was down to 5 minutes, which is a severe limitation that should be considered in the path planning and task assignment to correctly select the correct number of agents for achieving a full coverage of the infrastructure.

\subsection{System setup}
One of the most challenging issues when performing large scale infrastructure inspection is to keep a communication link with the agents performing the inspection, which is commonly used for monitoring the overall performance of the system. In this specific case, WiFi was the communication link of choice, mainly due to its simplicity of directly performing as expected, however it was quickly realized that the communication link was unstable due to height or occlusion of the MAV behind the wind turbine tower. To mitigate this issue, a different communication link should be used, e.g. the 4G cellular networks, and while WiFi can be used to upload mission trajectories it is not a reliable communication link at this scale.

Moreover, if it is desirable that the same mission can be executed again, the positions of the UWB anchors need to be kept. One possible way to achieve this is to consider the UWB anchors as supporting part of the infrastructure and have them permanently installed around the wind turbines, or to re-calibrate and consider the wind turbine as the origin, while only compensating for the rotation of the wind turbine depending on the mission setup.

\subsection{3D reconstruction}
Various visual sensors have been tested in the challenging case of wind turbine. The most beneficial sensor proved to be the monocular camera system. More specifically, the fixed baseline for stereo cameras can limit the depth perception and eventually degenerate the stereo to monocular perception. The reconstruction performance can also vary slightly, depending on the flying environment due to visual feature differences, therefore a robust and reliable, invariant to rotations feature tracker should be used. Another important factor for the reconstruction is the camera resolution, since it poses the trade off between higher accuracy and higher computational costs. Additionally, the path followed around the structure affects the resulting 3D model, which in combination with the camera resolution can vary the reconstruction results. Generally, the cameras should be calibrated and it is preferred to have set manual focus and exposure to maintain the camera parameters for the whole dataset. For SfM techniques it is required a large motion in rotation and depth among sequential frames to provide reliable motion estimation and reconstruction.

Moreover, a low cost LIDAR solution, that was tested during the field trials, failed to operate due to sunlight interfering with the range measurements. This sensor technology, should be further examined with more tests since they could be useful in obstacle avoidance and cross-section analysis algorithms.

\subsection{Localization}
While UWB positioning was the main localization system in the presented approach, it should be noted that this should not operate stand-alone. In the case of infrastructure inspection, one reference system should not act as a single point of failure, and it should be the aim to fuse as many sensors as possible. In the case of a wind turbine, the GPS does not provide a reliable position until the MAV is at significant height and the UWB localization system works best at lower height, hence it should be the aim to fuse both and utilize the sensor that is performing optimally depending on the current height. Moreover, neither UWB localization nor GPS provides a robust heading estimate, and the wind turbine causes magnetic disturbances that causing the magnetometers to fail and thus in this case visual inertial odometry is a robust solution to provide heading corrections since the landscape can be used as a stable attitude reference.

\section{Conclusions} \label{Conclusions}
%
This work presents a framework for autonomous visual inspection of a 3D infrastructure by utilizing multiple MAVs. To address this problem, the developed framework combined the fundamental tasks of path planning, localization and visual perception. Initially, a geometry-based path planner was employed for the collaborative coverage of complex structures, while the navigation of the platform has been performed through a localization component which provided accurate pose estimates of the MAVs by using a UWB-Inertial estimation scheme. Moreover, the inspection task considered compressed visual data streaming and visual data post processing for 3D model building. The performance of the proposed framework has the significant merit of being experimentally evaluated in realistic outdoor large scale infrastructure inspection experiments.
\section*{Acknowledgements} \label{Acknowledgements}
This work has received funding from the EU Horizon 2020 Research and Innovation Programme under the Grant Agreement No.644128, AEROWORKS. The authors would also like to thank Skellefte{\aa} Kraft for providing access to the windturbine site to perform the experimental trials.

\bibliography{mybib}

\begin{thebibliography}{10}

\bibitem{Kanellakis2017}
Christoforos Kanellakis and George Nikolakopoulos.
\newblock Survey on computer vision for uavs: Current developments and trends.
\newblock {\em Journal of Intelligent {\&} Robotic Systems}, pages 1--28, 2017.

\bibitem{mansouri2018cooperative}
Sina~Sharif Mansouri, Christoforos Kanellakis, Emil Fresk, Dariusz Kominiak,
  and George Nikolakopoulos.
\newblock Cooperative coverage path planning for visual inspection.
\newblock {\em Control Engineering Practice}, 74:118--131, 2018.

\bibitem{TORRES2016441}
Marina Torres, David~A. Pelta, José~L. Verdegay, and Juan~C. Torres.
\newblock Coverage path planning with unmanned aerial vehicles for 3d terrain
  reconstruction.
\newblock {\em Expert Systems with Applications}, 55:441 -- 451, 2016.

\bibitem{michael2012collaborative}
Nathan Michael, Shaojie Shen, Kartik Mohta, Yash Mulgaonkar, Vijay Kumar, Keiji
  Nagatani, Yoshito Okada, Seiga Kiribayashi, Kazuki Otake, Kazuya Yoshida,
  et~al.
\newblock Collaborative mapping of an earthquake-damaged building via ground
  and aerial robots.
\newblock {\em Journal of Field Robotics}, 29(5):832--841, 2012.

\bibitem{tomic2012toward}
Teodor Tomic, Korbinian Schmid, Philipp Lutz, Andreas Domel, Michael Kassecker,
  Elmar Mair, Iris~Lynne Grixa, Felix Ruess, Michael Suppa, and Darius
  Burschka.
\newblock Toward a fully autonomous uav: Research platform for indoor and
  outdoor urban search and rescue.
\newblock {\em IEEE robotics \& automation magazine}, 19(3):46--56, 2012.

\bibitem{sesar2016european}
JU~SESAR.
\newblock European drones outlook study.
\newblock {\em Unlocking the value for Europe. SESAR Joint Undertaking}, 2016.

\bibitem{perez2018architecture}
Francisco~J Perez-Grau, Ricardo Ragel, Fernando Caballero, Antidio Viguria, and
  Anibal Ollero.
\newblock An architecture for robust uav navigation in gps-denied areas.
\newblock {\em Journal of Field Robotics}, 35(1):121--145, 2018.

\bibitem{achtelik2014motion}
Markus~W Achtelik, Simon Lynen, Stephan Weiss, Margarita Chli, and Roland
  Siegwart.
\newblock Motion-and uncertainty-aware path planning for micro aerial vehicles.
\newblock {\em Journal of Field Robotics}, 31(4):676--698, 2014.

\bibitem{scaramuzza2014vision}
Davide Scaramuzza, Michael~C Achtelik, Lefteris Doitsidis, Fraundorfer
  Friedrich, Elias Kosmatopoulos, Agostino Martinelli, Markus~W Achtelik,
  Margarita Chli, Savvas Chatzichristofis, Laurent Kneip, et~al.
\newblock Vision-controlled micro flying robots: from system design to
  autonomous navigation and mapping in gps-denied environments.
\newblock {\em IEEE Robotics \& Automation Magazine}, 21(3):26--40, 2014.

\bibitem{GARCIAPULIDO2017152}
J.A. García-Pulido, G.~Pajares, S.~Dormido, and J.M. de~la Cruz.
\newblock Recognition of a landing platform for unmanned aerial vehicles by
  using computer vision-based techniques.
\newblock {\em Expert Systems with Applications}, 76:152 -- 165, 2017.

\bibitem{lupashin2014platform}
Sergei Lupashin, Markus Hehn, Mark~W Mueller, Angela~P Schoellig, Michael
  Sherback, and Raffaello D’Andrea.
\newblock A platform for aerial robotics research and demonstration: The flying
  machine arena.
\newblock {\em Mechatronics}, 24(1):41--54, 2014.

\bibitem{teixeira2017real}
Lucas Teixeira and Margarita Chli.
\newblock Real-time local 3d reconstruction for aerial inspection using
  superpixel expansion.
\newblock In {\em Robotics and Automation (ICRA), 2017 IEEE International
  Conference on}, pages 4560--4567. IEEE, 2017.

\bibitem{forster2015continuous}
Christian Forster, Matthias Faessler, Flavio Fontana, Manuel Werlberger, and
  Davide Scaramuzza.
\newblock Continuous on-board monocular-vision-based elevation mapping applied
  to autonomous landing of micro aerial vehicles.
\newblock In {\em Robotics and Automation (ICRA), 2015 IEEE International
  Conference on}, pages 111--118. IEEE, 2015.

\bibitem{fresk2017uwb}
Emil Fresk, Kristoffer {\"O}dmark, and George Nikolakopoulos.
\newblock Ultra wideband enabled inertial odometry for generic localization.
\newblock {\em IFAC-PapersOnLine}, 50(1):11465--11472, 2017.

\bibitem{schonberger2016structure}
Johannes~L Schonberger and Jan-Michael Frahm.
\newblock Structure-from-motion revisited.
\newblock In {\em Proceedings of the IEEE Conference on Computer Vision and
  Pattern Recognition}, pages 4104--4113, 2016.

\bibitem{seitz2006comparison}
Steven~M Seitz, Brian Curless, James Diebel, Daniel Scharstein, and Richard
  Szeliski.
\newblock A comparison and evaluation of multi-view stereo reconstruction
  algorithms.
\newblock In {\em Computer vision and pattern recognition, 2006 IEEE Computer
  Society Conference on}, volume~1, pages 519--528. IEEE, 2006.

\bibitem{msf}
S~Lynen, M~Achtelik, S~Weiss, M~Chli, and R~Siegwart.
\newblock A robust and modular multi-sensor fusion approach applied to mav
  navigation.
\newblock In {\em Proc. of the IEEE/RSJ Conference on Intelligent Robots and
  Systems (IROS)}, 2013.

\bibitem{mav_linear_mpc_}
Mina Kamel, Thomas Stastny, Kostas Alexis, and Roland Siegwart.
\newblock {\em Model Predictive Control for Trajectory Tracking of Unmanned
  Aerial Vehicles Using Robot Operating System}, pages 3--39.
\newblock Springer International Publishing, Cham, 2017.

\bibitem{pizer1987adaptive}
Stephen~M Pizer, E~Philip Amburn, John~D Austin, Robert Cromartie, Ari
  Geselowitz, Trey Greer, Bart ter Haar~Romeny, John~B Zimmerman, and Karel
  Zuiderveld.
\newblock Adaptive histogram equalization and its variations.
\newblock {\em Computer vision, graphics, and image processing},
  39(3):355--368, 1987.

\end{thebibliography}
\end{document}